%% file: main.tex
\documentclass[runningheads]{llncs}
\usepackage{graphicx}
\usepackage{multirow}
\usepackage{makecell}
\usepackage{amsmath}
\usepackage{amsfonts}
\usepackage{mathrsfs}
\usepackage{hyperref}

\newcommand{\x}{\textbf{x}}
\newcommand{\y}{\textbf{y}}

\usepackage{colortbl}
\usepackage{hhline}

\begin{document}
\title{A Client-server Deep Federated Learning for Cross-domain Surgical Image Segmentation}

\author{Ronast Subedi\inst{5} \and
 Rebati Raman Gaire\inst{5} \and
Sharib Ali\inst{4}\and 
Anh Nguyen\inst{3} \and 
Danail Stoyanov \inst{2} \and 
Binod Bhattarai\inst{1}\thanks{Corresponding author}
} 

\authorrunning{Subedi et al.}
\titlerunning{A Client-server Deep Federated...}
\institute{
University of Aberdeen, UK \and 
University College London, UK \and
University of Liverpool, UK \and 
University of Leeds, UK \and 
NepAl Applied Mathematics and Informatics Institute (NAAMII), Nepal \\
\email{\{binod.bhattarai\}@abdn.ac.uk}}

\maketitle        
\begin{abstract}
This paper presents a solution to the cross-domain adaptation problem for 2D surgical image segmentation, explicitly considering the privacy protection of distributed datasets belonging to different centers. Deep learning architectures in medical image analysis necessitate extensive training data for better generalization. However, obtaining sufficient diagnostic and surgical data is still challenging, mainly due to the inherent cost of data curation and the need of experts for data annotation. Moreover, increased privacy and legal compliance concerns can make data sharing across clinical sites or regions difficult. Another ubiquitous challenge the medical datasets face is inevitable domain shifts among the collected data at the different centers. To this end, we propose a Client-server deep federated architecture for cross-domain adaptation. A server hosts a set of  immutable parameters common to both the source and target domains. The clients consist of the respective domain-specific parameters and make requests to the server while learning their parameters and inferencing. We evaluate our framework in two benchmark datasets, demonstrating applicability in computer-assisted interventions for endoscopic polyp segmentation and diagnostic skin lesion detection and analysis. Our extensive quantitative and qualitative experiments demonstrate the superiority of the proposed method compared to competitive baseline and state-of-the-art methods. Codes are available at:
\url{https://github.com/thetna/distributed-da}.

\keywords{Domain Adaptation  \and
Federated Learning \and
Decentralised Storage \and Privacy}
\end{abstract}

\input{introduction}
\input{method}

\input{experiments}
\input{conclusions}
\clearpage
\bibliographystyle{splncs04}

\appendix
\section{Appendix}
\input{appendix}

\end{document}

%% file: introduction.tex
\section{Introduction}
The deployment of artificial intelligence (AI) technology in medical image analysis is rapidly growing, and training robust deep network architectures demands millions of annotated examples. Despite significant progress in establishing large-scale medical datasets, these are still limited in some clinical indications, especially in surgical data science and computer-assisted interventions~\cite{maier2022surgical}. Scaling training data needs multi-site collaboration and data sharing~\cite{ali2023multi}, which can be complex due to regulatory requirements (e.g. the EU General Data Protection Regulation~\cite{voigt2017eu}, and China's cyber power~\cite{inkster2018china}), privacy, and legal concerns. Additionally, even after training, practical AI model deployment in the clinic will require fine-tuning or optimization to local conditions and updates~\cite{hu2020challenges}. Therefore, architectures trained in federated and distributed ways to tackle cross-domain adaptation problems are critical. Yet, developing such architectures has challenges ~\cite{rieke2020future}. 

\begin{figure*}[t!]
    \centering
    \includegraphics[width=0.98\textwidth]{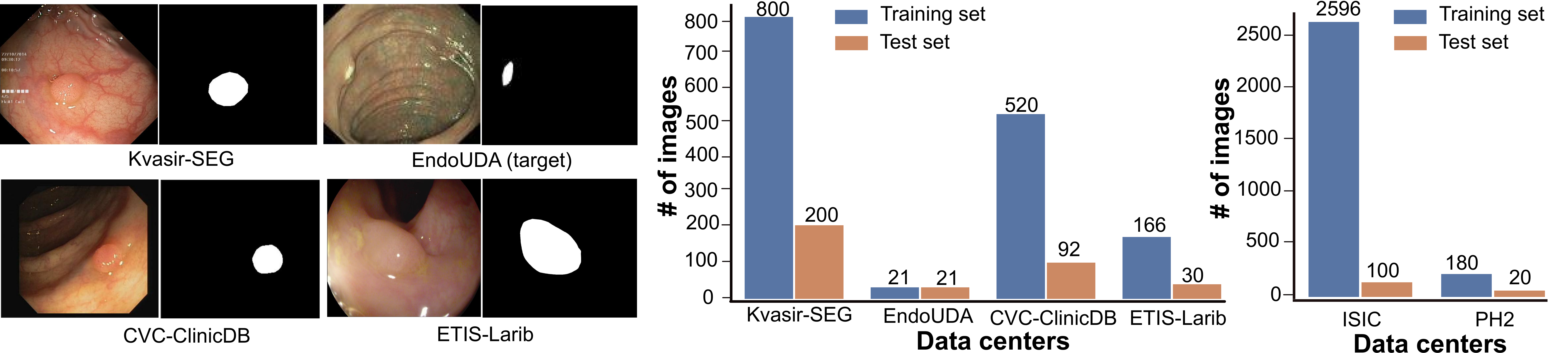}
    \caption{Sample training examples collected from various centres for polyp segmentation (left); Sizes of training and test set at different centres for polyp segmentation (middle) and  skin lesion segmentation (right).}
   
    \label{fig:introduction}
\end{figure*}

Several works~\cite{swati2019brain,ghafoorian2017transfer,gholami2019novel} have been proposed to tackle the problem of cross-domain adaptation in medical imaging. However, these methods require raw source and target domain data and cannot address the ever-increasing privacy concerns in sharing medical data. To circumvent the problem of privacy protection, there is a lot of research interest growing in Federated Learning (FL) in the medical domain~\cite{rieke2020future,li2020federated,sheller2020federated,tramel2020siloed,shen2022cd2,karargyris2021medperf,parekh2021cross}.Some methods even rely on synthetic data~\cite{hu2022fedsynth} to avoid sharing real data. For more details, we refer readers to a survey~\cite{nguyen2022federated} on federated learning for smart health care. The common drawback of most existing methods~\cite{rieke2020future,li2020federated,karargyris2021medperf,parekh2021cross} is that these methods are not designed for the domain shift problem. The most common topology in the FL workflow is averaging the local gradients (FedAvg) at the center and peer-to-peer gradient (FedP2P) sharing. These architectures are effective when data are independent and identically distributed (IID) in every client. In reality, domain shift is quite prevalent as data collected at different centers tend to be center specific. In Fig.~\ref{fig:introduction}, we can see the training examples for polyp segmentation collected at different centres. These examples show the discrepancy in lighting, camera pose and modalities in different centers. Some recent works, such as by Guo et al.~\cite{guo2021multi} and FedDG by Liu et al.~\cite{liu2021feddg}, address cross-domain problems in FL. However,~\cite{guo2021multi} limits to a source-target pair at a time. Also, they employed adversarial loss to align the parameters, which is difficult to optimize. Similarly, FedDG~\cite{liu2021feddg} shares the information between the sources in the form of amplitudes of images. Their evaluation is limited to fundus images and MRI. 

To tackle the problems of cross-domain adaptation and privacy protection in surgical image segmentation, we propose a simple yet effective Client-server FL architecture consisting of a server and multiple clients, as shown in Fig.~\ref{fig:proposed_arch}. A server hosts a set of \emph{immutable} task-specific parameters common to all the clients. Whereas every client requests the server to learn their domain-specific parameters locally and make the inference. In particular, every client learns an encoder's parameters to obtain an image's latent representation. These latent representations and ground truth masks are sent to the server. The decoder deployed on the server makes the predictions and computes the loss. The gradients are computed and updated only on the encoder to align the client's features with task-specific parameters hosted on the server. Aligning domain-specific parameters to common parameters helps diminish the gap between the source and target domains. We can draw an analogy between our framework with public-key cryptography.  A client's network parameter is equivalent to a private key, and the decoder's parameters shared on the server are equivalent to the public key. Thus a client only with access to its private key can transfer its latent vector to the server containing the public key to obtain the semantic mask. Distributed storage of the parameters diminishes the risk of model parameter theft and adversarial attacks~\cite{ma2021understanding}. Moreover, each client communicates to the server only via a latent image representation, which prevents exposing the information of the raw data collected on the client side. It is possible to encrypt data transfer between the server and clients to make it more secure. Finally, the server receives only fixed latent dimension representations, making it agnostic to the client's architecture. This enables clients to communicate with the server concurrently, improving efficiency. Likewise, none of the centres can modify the parameters deployed on the server; this would prevent the memorisation of client-specific information and parameter poisoning on the server~\cite{kurita2020weight}.

To sum up, we propose a Client-server Federated Learning method for cross-domain surgical image segmentation. We applied our method to two multi-centre datasets for endoscopic polyp and skin lesion segmentation. We compared with multiple baselines, including recent works on cross-domain FL~\cite{liu2021feddg,guo2021multi} and obtained a superior performance. 

%% file: method.tex
\section{Method}
\label{sec:method}
\noindent \textbf{Background:} 
We consider a scenario where we have $C_1, C_2, \dots C_n$ represent $n$ number of  different institution's centres located at various geographical regions. Each centre collects its data in the form of tuple  $(\x, \y)$
 where $\x \in \mathbb{R}^{w \times h \times c}, \y \in \mathbb{R}^{w \times h}$, where, $w, h, c$ represent the width, height, and number of channels of an image. The annotated examples collected at the different centers are not IID due to variations in the illumination, the instruments used to acquire data, the ethnicity of the patients, the expertise of the clinician who collects the data, etc. We denote the total number of annotated pairs in each centre by $N_n$. In this paper, one of the major goals is to address the problem of domain adaptation, avoiding the need for the sharing of raw data to protect privacy. 

\begin{figure}
    \centering
   \includegraphics[width=0.98\linewidth]{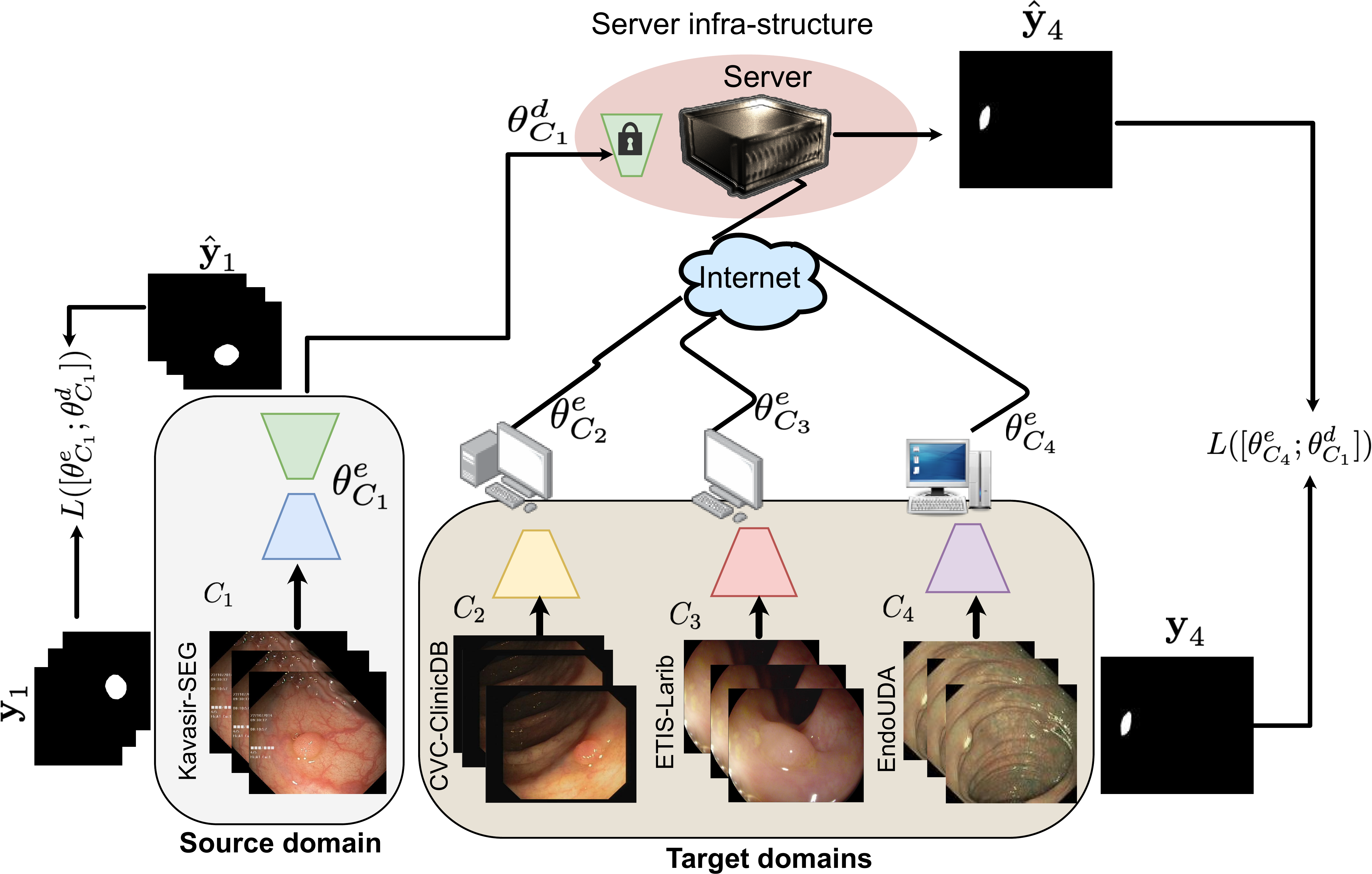}
    \caption{The schematic diagram of the proposed framework. There are three major components: Source domain, Target domains, and Server infrastructure to share common parameters.}
    \label{fig:proposed_arch}
\end{figure}

\noindent \textbf{Learning Source Domain Parameters:} 
First, we train  a semantic segmentation network  on the source data. In Figure~\ref{fig:proposed_arch}, the Source domain block shows the training of source domain/centre parameters. For us, data collected on Centre 1, $C_1$  is source data. We employ fully-convolutional encoder-decoder architecture. Such architectures
are quite popular for semantic segmentation~\cite{badrinarayanan2017segnet,ronneberger2015u}. With the randomly initialised parameters, we minimise the objective of the Equation~\ref{eqn:semantic_seg}. In Equation~\ref{eqn:semantic_seg}, $\theta_{C_1}^{e}$ and $\theta_{C_1}^{d}$ represent the learnable parameters of the encoder and decoder, respectively.   

 \begin{equation}
 \begin{split}
 L([\theta_{C_1}^{e}; \theta_{C_1}^{d}]) = -\frac{1}{N_1} \sum_{i=1}^{N_1} \sum_{i=j}^{W}  \sum_{k=1}^{H} \y_{ijk}\log\hat{\y}_{ijk}
 \\ 
 + (1 - \y_{ijk})\log(1-\hat{\y}_{ijk}) 
 \end{split} 
 \label{eqn:semantic_seg}
 \end{equation} 

\noindent \textbf{Setting-up Server Infra-structure:} 
Figure~\ref{fig:proposed_arch} Server infra-structure block shows the setting up of the server infrastructure. 
Once we learn the parameters of the network from the source ($C_1$) data set, we upload the decoder ($\theta_{C_1}^{d}$) parameters on the server to share with every target client.  The decoder module specializes to segment anatomies, given the encoder module's latent vector representation of the input image. As this segmentation task is common to all the centres, we propose to use a single decoder for all the centres. The previous works on cross-modal~\cite{castrejon2016learning} and cross-feature~\cite{bhattarai2016deep} representations learning for cross-domain classification in computer vision employed the idea of sharing Convolutional Neural Networks' top layers parameters. However, none of these methods were employed in Federated Learning. The idea of sharing top layers parameters is in contrast to the conventional transfer learning~\cite{huh2016makes} where the parameters are initialised with a model pre-trained on Imagenet and fine-tuned only the task-specific fully connected layers. We freeze the shared decoder parameters of the source. 
This arrangement brings advantages for privacy protection by preventing weight poisoning attacks~\cite{kurita2020weight}. Weight poisoning attacks alter the pre-trained model's parameters to create a backdoor. The parameters of the encoders can be shipped to the target client as per demand. Sometimes, the clients may demand these parameters to initialize their local networks when the training data is very small. 

\noindent \textbf{Federated Cross-domain Adaptation:} 
Target centres other than the source centre deploy only the encoder network of their choice.
In Figure~\ref{fig:proposed_arch}, the target domain block depicts it. Every centre feeds its images to its encoder network during training, generating the respective latent representations. The latent representation and the ground truth ($\y_i$) mask from each target centre are pushed to the server where the pre-trained source decoder, $\theta_{C1}^{d}$, is placed. The decoder feeds the latent representation, which predicts the output segmentation labels ($\hat{\y_i}$). We learn the parameters of the target encoders ($\theta^{e}_{Ci}$) to minimize the objective given in Equation~\ref{eqn:semantic_seg_client}. Since the decoder parameters are frozen and shared with every client, only the target encoder's parameters are updated on the client side. This helps to align the latent representations to that of the source decoder's parameters and maximises the  benefit from the task-specific discriminative representations learned from the large volume of source data.  
\vspace{-0.7cm}
 \begin{equation}
 \begin{split}
 \\
 L([\theta_{C_i}^{e}; \theta_{C_1}^{d}]) = -\frac{1}{N_i} \sum_{i=2}^{N_i} \sum_{i=j}^{W}  \sum_{k=1}^{H} \y_{ijk}\log\hat{\y}_{ijk} \\ + (1 - \y_{ijk})\log(1-\hat{\y}_{ijk}) \\
 \forall i \in {2, \dots n}
 \end{split}
 \label{eqn:semantic_seg_client}
 \end{equation} 

The only thing that matters for target centres to communicate to the server is the fixed dimension of latent representations of an image. Thus, our architecture gives the flexibility of deploying the various sizes of networks on the client side based on available computing resources. And it is also entirely up to the target centres whether they want to initialize the parameters of the encoder using the parameters of the source domain. If the number of training examples is extremely few, then initialization using the pre-trained model’s weight can prevent over-fitting.

%% file: experiments.tex
\section{Experiments}
\label{sec:experiments}

\noindent \textbf{Data sets and Evaluation Protocol:} 
We applied our method in two benchmark datasets: endoscopic polyp segmentation and skin lesion segmentation. 
The \textbf{polyp segmentation dataset} contains images collected at four different centres. 
Kvasir-SEG~\cite{jha2020kvasir} data set makes the source centre ($C_1$) in our experiment. It has 800 images in the train set and 200 in the test set. These high-resolution images acquired by electromagnetic imaging systems are made available to the public by Simula Lab in Norway. Similarly, the EndoUDA-target data set makes the first target domain ($C_2$) in our experiment, consisting of 21 images in both the training and testing sets~\cite{celik2021endouda}. Our experiment's second target domain centre~($C_3$) consists of images from the CVC-ClinicDB dataset made available to the research community by a research team in Spain. There are 520 images in the train set and 92 in the test set. Finally, the ETIS-Larib dataset released by a laboratory in France makes our third target domain data set ($C_4$). This data set consists of 166 in the train set and  30 images in the test set. These data sets were curated at different time frames in different geographical locations. 

\begin{table*}
\small 
\centering
\begin{tabular}{ c|c|c|c|c|c|c|c|>{\columncolor[gray]{0.95}}c|>{\columncolor[gray]{0.85}}c} 
\multirow{2}{*}{Data} & \multirow{2}{*}{Centres} & \multicolumn{8}{|c}{mIOU} \\
\cline{3-4} \cline{5-10} 
\hhline{~|~|-|-|-|-|-|-|-|-}
  && INDP & COMB & {~\cite{konevcny2015federated}} & {\cite{liu2021feddg}} & FtDe &{\cite{guo2021multi}} &  RandEn & FtEn  \\
\hline
\multirow{4}{*}{Endo.}&  Kvasir-SEG ($C_1$, source) &  80.3 & \textcolor{blue}{81.0} & \textcolor{red}{82.3} & 73.5 &N/A & 80.5 & 80.3 & 80.3   \\ 
 &  EndoUDA ($C_2$) & 52.0 & 57.5& 53.1 & 29.7 &59.9 & \textcolor{blue}{61.7} & 50.6 &\textcolor{red}{62.0}  \\
& CVC-ClinicDB ($C_3$) & 88.3 & 87.8 & 86.8 & 74.5 & 85.8 & 83.0 & \textcolor{red}{89.1} & \textcolor{blue}{88.4} \\ 
&  ETIS-Larib ($C_4$) & 62.1 & 66.9 & 61.4 & \textcolor{blue}{70.8} & 65.1 & \textcolor{red}{71.7} & 64.3 & 69.9 \\
\hline
\multirow{2}{*}{Skin} & ISIC ($C_1$, source) & 81.3 & 75.7 & \textcolor{red}{84.9} & N/A& N/A & NA& 81.3 & N/A \\ 
 & PH2 ($C_2$) & 88.4 & 88.3 & 88.4 & N/A& 88.0 &NA  & \textcolor{red}{89.6} & 89.4  \\ 
\hline 
\end{tabular}
\label{tab:quant_polyp_seg}
\caption{mIoU scores on Endoscopic Polyp Segmentation Data sets (upper block) and Skin Lesion Segmentation (lower block).  }
\end{table*}

For~\textbf{skin lesion segmentation}, we took data set collected at two different centres: 
ISIC(International Skin Imaging Collaboration)~\cite{codella2019skin} and PH2~\cite{mendoncca2013ph}. 
In ISIC, there are 2596 training examples and 102 test examples. The PH2 database is curated through a joint research collaboration between the Universidade do Porto, Tecnico Lisboa, and the Dermatology Service of Hospital Pedro Hispano in Matosinhos, Portugal. In this data set, there are only 180 training examples and 20 testing examples. We consider ISIC and PH2 source and target domain, respectively. We use the mean Intersection of Union (mIoU) and dice score for both datasets for quantitative evaluations. Qualitative comparisons also validate our idea.

\noindent \textbf{Baselines:}
We have compared the performance of our method with several competitive baselines, including both non-federated and federated frameworks. One of the naive baselines is to train a model for each target centre independently (\emph{INDP}). The models of the centres with less training data overfit. Another configuration is creating a data pool by combining the training data (\emph{COMB}) from all the centres and training a single model. However, this method does not address any of the issues regarding privacy and compliance. Another viable option is to adapt a pre-trained model to a new domain by fine-tuning the parameters of the latter layers (\emph{FtDe}). We also compared our method with competitive federated learning algorithms. FedAvg~\cite{konevcny2015federated} averages the gradients computed in every center and shares the average gradients with the clients. This method ignores the non-IID nature of data from different centres. FedDG~\cite{liu2021feddg} is another Federated Learning method for domain adaption published at CVPR 2021. Finally, we also compared with another recent work by Guo et al.~\cite{guo2021multi} for federated learning for multi-institutional data published at CVPR 2021. Our methods have two variants: initialising clients' parameters randomly(\textbf{RandEn}) and with the source's parameters (\textbf{FtEn}) 

\begin{figure}
    \centering
    \includegraphics[width=0.75\textwidth]{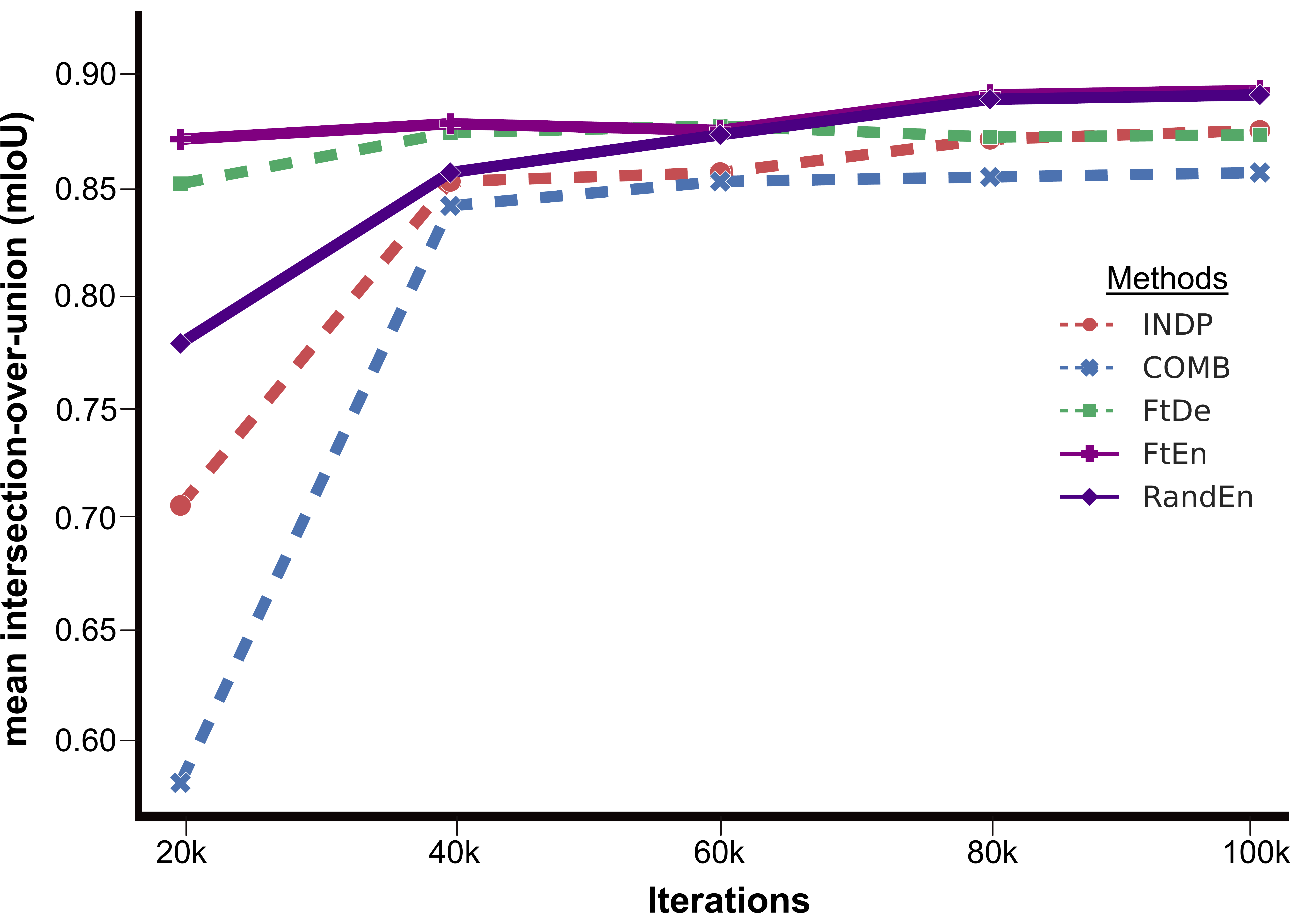}
    \caption{Curves show the learning behaviour of different methods on PH2 data set.}
    \label{fig:learning_curve_skin}
\end{figure}

\noindent \textbf{Implementation Details and Learning Behaviour:}
We implement our algorithms on PyTorch framework. All the images were resized to the dimension of $418\times418$. 
For optimization, we employ Adam optimizer with values of $\beta1$ and $\beta2$ set to 0.9 and 0.999 respectively. We initialize learning rate to 2e-4 and set the decaying of the learning rate every 25k iterations. Taking UNet as a base architecture, we train the networks for 100k iterations and save the best-performing checkpoints on the validation set and report the performance on the test set. Figure~\ref{fig:learning_curve_skin} summarises the learning behaviour of the different methods for the first 100k iterations on PH2 data set, a target domain for skin lesion segmentation. The solid lines are our methods, and the dashed lines are the compared methods. The smooth curves demonstrate that our methods are easy to optimise the parameters.    

\noindent \textbf{Quantitative Evaluations:}
Table~\ref{tab:quant_polyp_seg} shows the quantitative performance comparison. In the table, the last two grey-shaded columns show the performance of our methods. Our method outperforms \emph{INDP} in every target centre. 
This signifies the importance of domain adaptation by our method. Compared to the other Federated Learning methods, our methods obtain the highest performance on 2/3 of target centres and are competitive on the third one for endoscopic polyp segmentation. On skin lesion segmentation, our method surpassed all the compared baselines and the recent competitive Federated Learning methods. We have compared the performance on the dice score, too and obtained a similar performance (See Supplementary). We have also evaluated the performance with varying sizes of the target domain/centre encoders. Please see the appendix for the details.

\begin{figure*}
    \centering
    \includegraphics[width=0.95\textwidth]{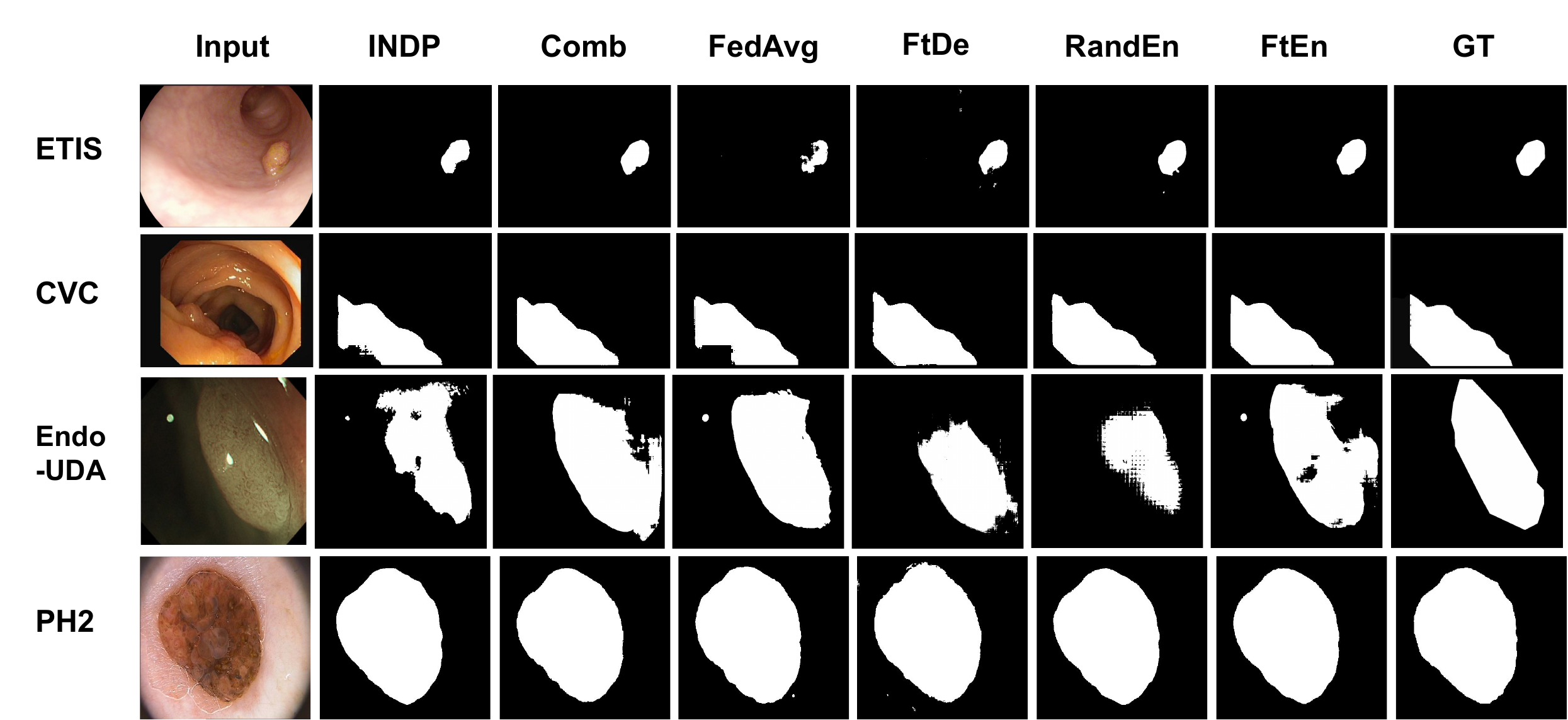}
    \caption{Qualitative Comparisons}
    \label{fig:qualitative_results}
\end{figure*}

\noindent \textbf{Qualitative Evaluations:}
Figure~\ref{fig:qualitative_results}  shows the qualitative performance comparisons between the baselines and 
the proposed methods on the target domains. Rows 2-4 (inclusive) are from endoscopy benchmarks, and the last row is from skin benchmarks. FedAvg, our closest work, fails to generalise well on target domains (see ETIS-Larib and CVC-ClinicDB). Whereas our method is consistent in every target domain. These results further validate that our method is superior to the others. You can find more examples in the appendix section.

\noindent \textbf{Computational Complexity:}  
 For INDP and FedAvg, the parameters of both the encoders and decoders grow in O(n)  with the number of centres. Similarly, for FtDe and our method, the parameters of the encoder grow in O(n), while the parameters of the decoder are constant, i.e., O(1). Although the growth of the parameters for both the encoder and decoder for COMB is O(1), it does not address any privacy concerns. From these, our method is computationally less expensive and has high privacy protection. 

%% file: conclusions.tex
\section{Conclusions}
In this paper, we presented a client-server Federated Learning architecture for cross-domain surgical image segmentation. Our architecture addresses the cross-domain adaptation problem without sharing the raw images. Moreover, sharing only a part of the parameters from the source domain enhances privacy protection. Extensive experiments on two benchmarks from various data centres demonstrated improved cross-domain generalisation and privacy protection over the baselines and the competitive contemporary method.

\section{Acknowledgements}
This work is partly supported by the Wellcome/EPSRC Centre for Interventional and Surgical Sciences (WEISS) [203145Z/16/Z]; Engineering and Physical Sciences Research Council (EPSRC) [EP/P027938/1, EP/R004080/1, EP/P012841/1]; The Royal Academy of Engineering Chair in Emerging Technologies scheme; and the EndoMapper project by Horizon 2020 FET (GA 863146).

%% file: appendix.tex
\noindent \textbf{Varying Encoder Size:}
To further validate the robustness of our proposed framework, we trained the networks with different numbers of learnable parameters in the encoder module as shown in table~\ref{tab:quant_varying_encoder_size}. We vary the parameters by adding or removing the constituting layers in the encoding blocks of the network. We designate the conventional encoder of the UNet architecture as an encoder with medium size. The learnable parameters in the medium encoder are approximately 17 million. Table~\ref{tab:general_enc_architecture} depicts the architecture of a general encoder. It consists of three down-sampling layers, represented as \textit{Down Block}.  We vary the number of layers in the Down Blocks of the general encoder as per the availability of labeled data and computing resources in the particular centre.
\begin{table}
    \centering
    \begin{tabular}{c|c|c|c||c|c}
     \thead{\multirow{2}{*}{Data}} & \thead{\multirow{2}{*}{Centres}}  & \thead{\multirow{2}{*}{Size}}  &
     \multirow{2}{*}{\thead{Trainable \\ Parameters} } &
     \multicolumn{2}{|c}{\thead{mIoU}}  \\
     \cline{5-6}
     \rule{0pt}{3ex}&&&&\thead{INDP}&\thead{RandEnc(Ours)} \\
     \hline
     \multirow{3}{*}{Polyp}& \multirow{3}{*}{ETIS-Larib}& Small&6,311,616 &61.6&62.9 \\
      \cline{3-6}  
     & & Medium& 17,080,896 &62.1&64.3 \\
      \cline{3-6}  
      && Large& 27,850,176&62.5& 64.7\\
    \hline
     \multirow{3}{*}{Skin}& \multirow{3}{*}{PH2}& Small&6,311,616 &88.4&88.4 \\
      \cline{3-6}  
     & & Medium& 17,080,896 &88.5&89.6 \\
      \cline{3-6}  
      && Large& 27,850,176&89.4& 89.8\\
    \hline
    \end{tabular}
    \caption{Result on Endoscopic Polyp Segmentation Data sets (upper block) and Skin Lesion Segmentation (lower block) with different sizes of encoder networks.}
    \label{tab:quant_varying_encoder_size}
\end{table}
\vspace{-4em}
\begin{table}[h]
        \centering
        \begin{tabular}{ c c } 
        \hline
        \rule{0pt}{2.2ex}\textbf{Operations} & \textbf{Output Size} \\[0.5ex]
        \hline
        \rule{0pt}{1.3ex} Input Image & $c \times h \times h$ \\[0.5ex]
        \hline
        \rule{0pt}{1.3ex}Down Block & $c\times \frac{h}{2} \times \frac{h}{2}$ \\[0.5ex]
        \hline
        \rule{0pt}{1.3ex}Down Block & $c\times \frac{h}{4} \times \frac{h}{4}$ \\[0.5ex]
        \hline
        \rule{0pt}{1.3ex}Down Block & $c\times \frac{h}{8} \times \frac{h}{8}$ \\[0.5ex]
        \hline
        \end{tabular}
       \caption{Architecture of the general encoder.}
        \label{tab:general_enc_architecture}
\end{table}
\vspace{-2em}

The proposed federated distributed framework for domain adaptation provides flexibility to choose different network architectures to learn a common latent representation of images. These architectures can be designed based on the data size and available resources at a particular center. The performance evaluation results of various encoder sizes on polyp segmentation and skin lesion segmentation are shown in table~\ref{tab:quant_varying_encoder_size}. From the table, we can observe that segmentation performance for various sizes of the encoder is higher when trained using our framework than training independently for a specific center.\newline

\noindent\textbf{Equations for Gradient Update:} Equations~\ref{eqn:grad_update_dec} and~\ref{eqn:grad_update_enc} show mathematical formulation to update gradients of decoder and encoder modules in the proposed framework. Equation~\ref{eqn:grad_update_dec} updates the gradient of the decoder in the source center. Equation~\ref{eqn:grad_update_enc} represents the mechanism of updating gradients of each encoder module in every center. Here, $\alpha$ is the center-specific learning rate. Please note the parameters of the decoder remain the same for every target centre. $L$ denotes the loss function.

\begin{equation}
    \theta_{C_1}^{d} = \theta_{C_1}^{d} - \alpha_{1} \times \frac{ \partial L([\theta_{C_1}^{e}; \theta_{C_1}^{d}])}{\partial \theta_{C_1}^{d}};
    \label{eqn:grad_update_dec}
\end{equation}

\begin{equation}
    \theta_{C_i}^{e} = \theta_{C_i}^{e} - \alpha_{i} \times \frac{ \partial L([\theta_{C_i}^{e}; \theta_{C_1}^{d}])}{\partial \theta_{C_i}^{e}} ; \forall i \in {1, \dots n}
    \label{eqn:grad_update_enc}
\end{equation}

\begin{figure}
    \centering
    \includegraphics[trim= 0cm 7cm 0cm 0.8cm, clip, width=0.95\textwidth]{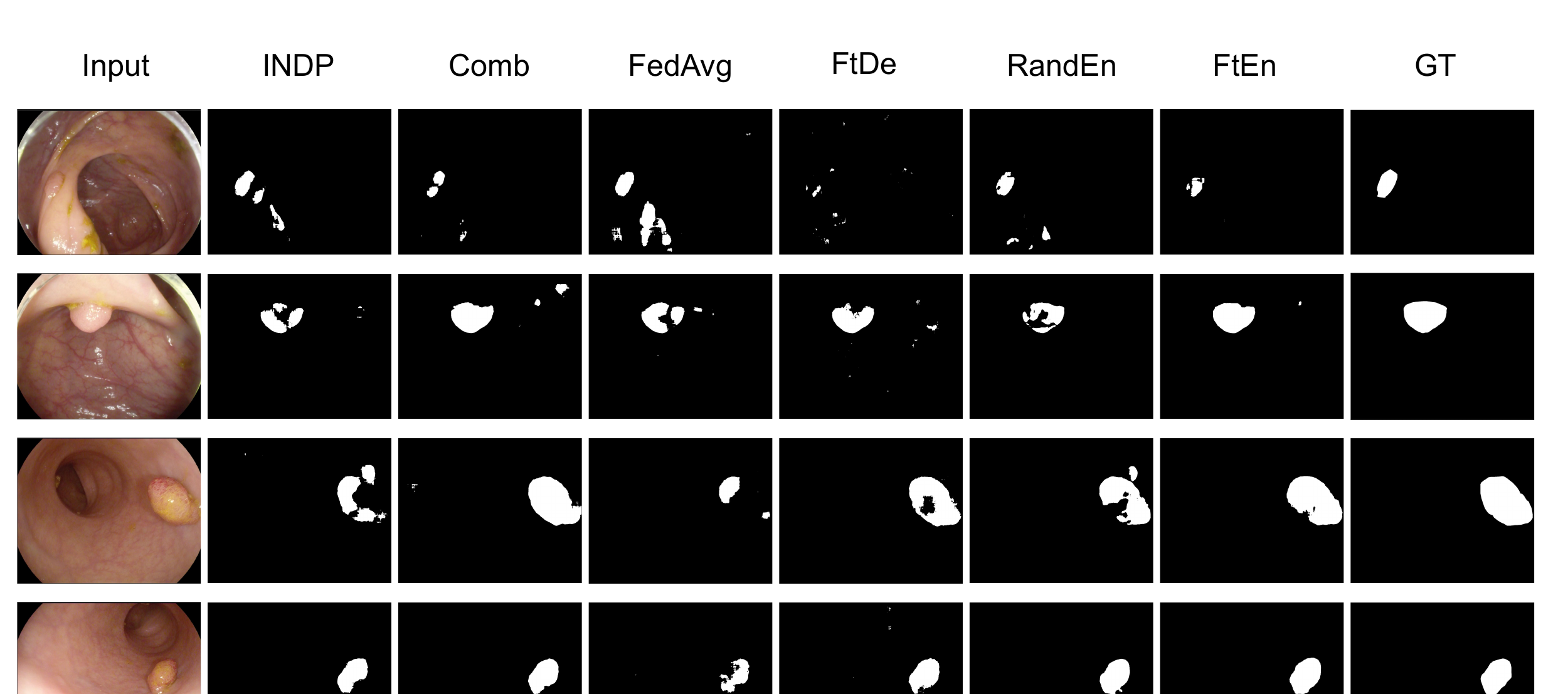}
    \includegraphics[trim= 0cm 7cm 0cm 5cm, clip, width=0.95\textwidth]{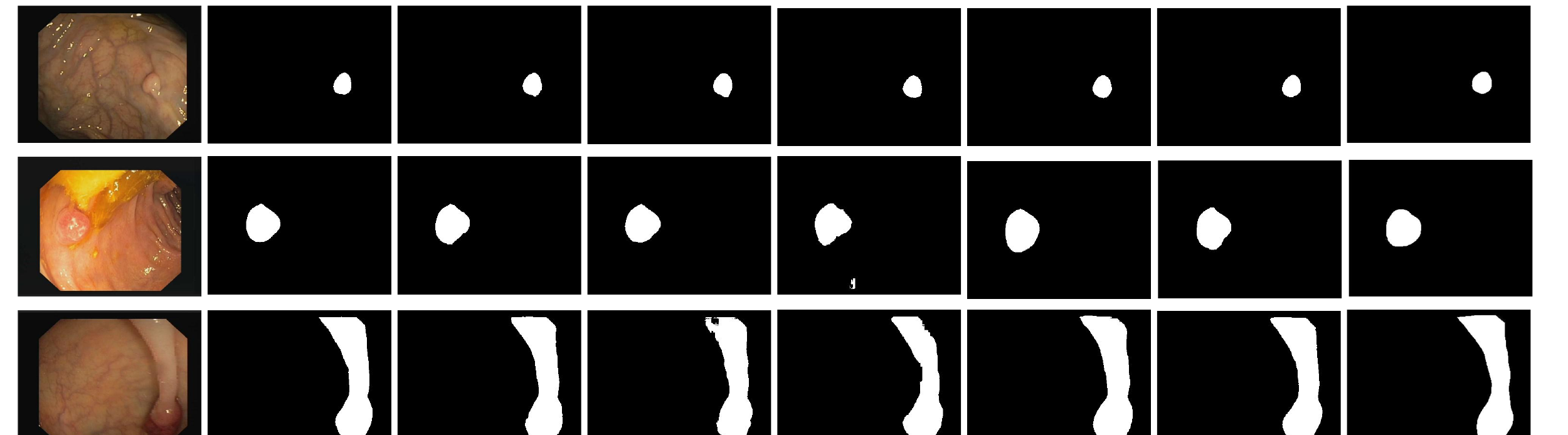}
    \includegraphics[trim= 0.25cm 12cm 1.5cm 2.2cm, clip, width=0.95\textwidth]{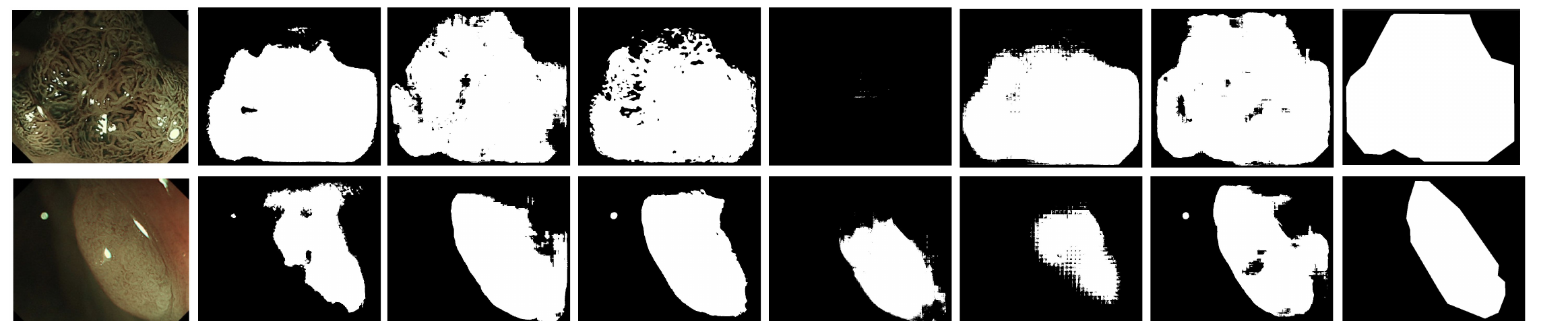}
    \vspace{-0.5em}
    \caption{Additional Qualitative Comparison}
    \label{fig:add_qualitative}
\end{figure}

\noindent\textbf{Additional Qualitative Results:} Figure~\ref{fig:add_qualitative} depicts the segmentation results on Endoscopic polyp data from the target domain centres. The first four rows are from the ETIS-Larib dataset, the next three are from the CVC-ClinicDB data set, and the last two are from EndoUDA. Columns \textbf{RandEn} and \textbf{FtEn} show the results of our method.
\vspace{-2.25em}

